\documentclass[sigconf]{acmart}

\AtBeginDocument{%
  }

\renewcommand\footnotetextcopyrightpermission[1]{}

\begin{document}
\pagestyle{plain}

\title{PL-NBA: A Possession-level Universal Basketball Video Dataset Supporting Multiple Visual Understanding Tasks}

\author{Yunhao Zhao}
\email{holhouse@emails.bjut.edu.cn}
\orcid{0009-0005-7036-535X}
\affiliation{%
  \institution{Beijing University of Technology}
  \city{Beijing}
  \country{China}
}

\author{Haoying Sun}
\email{sunhaoying97@163.com}
\affiliation{%
  \institution{Beijing University of Technology}
  \city{Beijing}
  \country{China}}

\author{Jiarui Li}
\email{ljr2024@emails.bjut.edu.cn}
\affiliation{%
  \institution{Beijing University of Technology}
  \city{Beijing}
  \country{China}
}

\author{Zhuming Wang}
\email{wzm1030@126.com}
\affiliation{%
 \institution{Beijing University of Technology}
 \city{Beijing}
 \country{China}
 }

\author{Ya Jing}
\email{jingya@bjut.edu.cn}
\affiliation{%
 \institution{Beijing University of Technology}
 \city{Beijing}
 \country{China}
 }

\author{Xiangbo Shu}
\affiliation{%
 \institution{Nanjing University of Science and Technology}
 \city{Nanjing}
 \country{China}
 }

\author{Lifang Wu}
\authornote{Corresponding Author} 
\email{lfwu@bjut.edu.cn}
\affiliation{%
 \institution{Beijing University of Technology}
 \city{Beijing}
 \country{China}
 }

\author{Changwen Chen}
\email{changwen.chen@polyu.edu.hk}
\affiliation{%
 \institution{Hong Kong Polytechnic University}
 \city{Hong Kong}
 \country{China}
 }

\begin{abstract}
Visual understanding in sports has emerged as a hot topic in computer vision in recent years. Most existing basketball video datasets adopt single action or activity as sample, which can neither preserve the temporal continuity of game events nor support complex tasks such as action anticipation. To address this issue, this paper constructs the first possession-level basketball video dataset (PL-NBA), in which each sample is composed of a complete NBA offensive possession. Collected from 60 NBA games, PL-NBA contains  11,000 valid offensive possession clips and 31,567 annotated events with player names, captions, event types and timestamps. Each video clip includes multiple events and preserves the continuity of events, which is helpful for analysis of tactic. Experiment is conducted on multiple visual understanding tasks, including event recognition, video captioning, temporal action localization and action anticipation. Experimental results show that existing methods achieve limited performance on above four tasks, demonstrating that PL-NBA is a challenging benchmark for sports video understanding.
\end{abstract}

\keywords{Basketball video dataset, Visual Understanding, Event recognition}

\begin{teaserfigure}
  \centering
  \scalebox{0.9}{
    \includegraphics[width=0.85\textwidth]{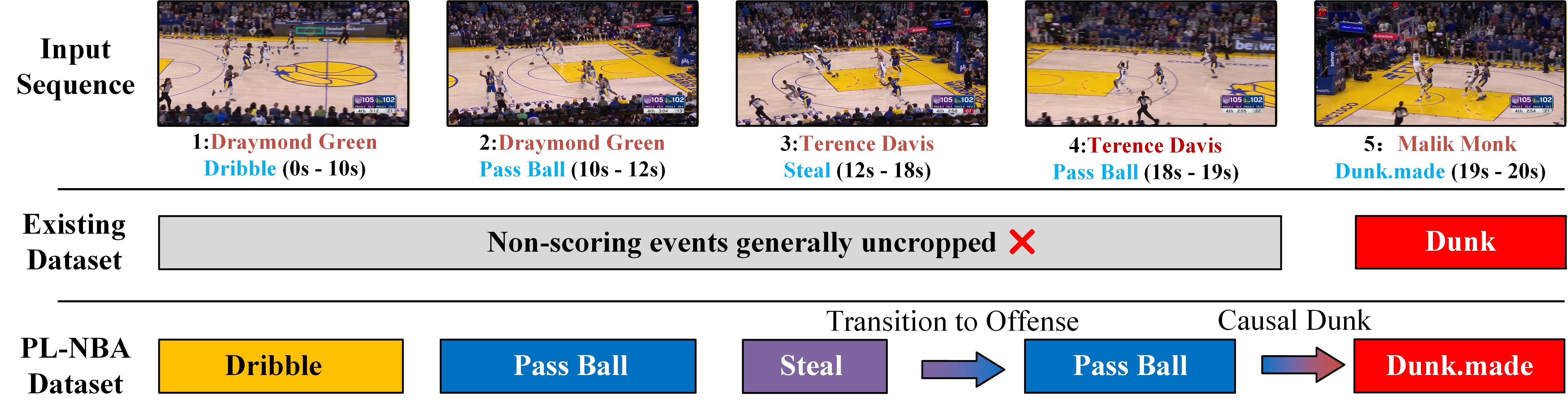}
  }
  \caption{Existing datasets (middle) primarily annotate isolated event clips, overlooking the inherent temporal dynamics of basketball games. In contrast, our proposed PL-NBA dataset (bottom) captures all constituent events within a complete offensive possession. This provides unified, high-quality data support for various downstream tasks.}
  \label{fig1}
\end{teaserfigure}

\maketitle

\begin{table*}
  \caption{Comparison between existing basketball video datasets and our PL-NBA dataset.}
  \label{tab1}
  \centering
  \begin{tabular}{@{}lccccccc@{}}
    \toprule
    Name & FSN & NBA & NSVA & BH-Commentary & VC-2022 & FineSports & PL-NBA  \\
    \midrule
    Year & 2018 & 2020 & 2022 & 2024 & 2024 & 2024 & 2026 \\
    \midrule
    Sample Level & event & event & event & event & event & event  & possession \\
    Number of Samples & 2,000 & 9,172 & 32,019 & 4,396 & 11,489 & 10,000 &  11,000 \\
    \midrule
    Data Source & NBA & NBA:18-19 & NBA:18-19 & NBA:20-23 & NBA:22-23 & NBA & NBA:22-25 \\
    Video Resolution & --- & 1080P & 720p & 720p & 720p & --- & 1080P \\
    Audio Track & × & × & × & × & × & × & \checkmark \\
    \midrule
    Event Recognition & × & \checkmark & \checkmark & × & × & × & \checkmark \\
    Temporal Action Localization & × & × & × & × & × & × & \checkmark \\
    Spatiotemporal Action Localization & × & × & × & × & × & \checkmark & × \\
    Video Captioning & \checkmark & × & \checkmark & \checkmark & \checkmark & × & \checkmark \\
    Action Anticipation & × & × & × & × & × & × & \checkmark \\
    \bottomrule
  \end{tabular}
\end{table*}

\section{Introduction}
Visual understanding tasks in sports constitute a pivotal interdisciplinary research area integrating computer vision and sports science. In recent years, significant progress has been made in numerous sub-tasks, such as video captioning\cite{FSN,xi2025player,vc1,xi2025eika}, event recognition\cite{ac1,ac2,ac3,ac4}, and group activity recognition\cite{gar1,gar2,gar3,gar4}.

However, existing sports video understanding research is constrained by specific datasets—relevant tasks focus mainly on the action level with single-dimensional annotations. Such datasets can only support single sub-tasks and fail to reflect the contextual dependencies between actions, thus failing to support complex application scenarios like tactical analysis and intelligent broadcast.

Among various sports disciplines, basketball, characterized by its fast pace and diverse types of action, has become a key research hotspot for visual understanding. In a basketball game, an offensive possession serves as the fundamental unit of tactical execution. It is widely adopted to calculate offensive efficiency and defensive efficiency, acting as a critical benchmark to evaluate team performance. Meanwhile, it also serves as the basic unit for the audience to understand the progression of game and tactical logic. A possession may include a series of coherent actions, such as dribbling, passing, and shooting. As illustrated in Figure~\ref{fig1}, there are clear logical connections between consecutive actions.

In summary, a possession-level basketball video dataset can not only support a simple visual understanding task but also adapt to various complex research needs, promising to advance basketball video understanding toward a more practical direction. To this end, this study constructs a new basketball game video dataset named PL-NBA. Unlike traditional datasets centered on individual events, each sample in PL-NBA contains a complete offensive possession, fully preserving the continuity of action sequences and the integrity of tactical scenarios.

\noindent \textbf{The contributions of this paper are as follows:}

(1) We construct the first possession-level basketball video dataset, named PL-NBA. It includes 11,000 valid offensive possessions with 31,567 events collected from 60 NBA games, and multiple labels including player names, captions, event types, and timestamps. It can provide comprehensive annotation support for various sports visual understanding tasks.

(2) We conduct benchmark experiments on event recognition, video captioning, and temporal action localization using PL-NBA, validating the effectiveness of our dataset.

(3) We propose the action anticipation task in basketball games based on the PL-NBA dataset and present a baseline method, which provides a novel task paradigm and a feasible baseline reference for the research on intelligent analysis of basketball events.
\begin{figure*}[t]
  \centering
  \includegraphics[width=0.9\linewidth]{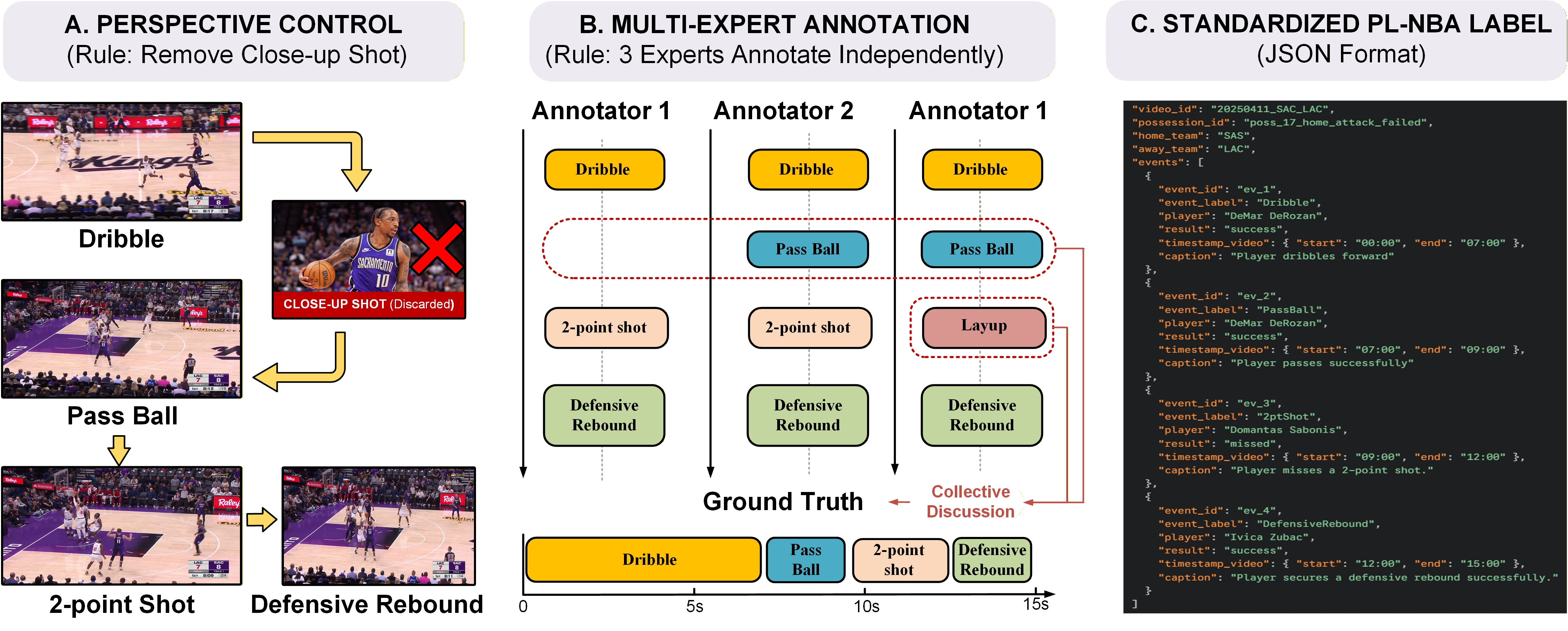}
   \caption{Overview of the PL-NBA annotation workflow. The pipeline ensures data quality through perspective control (retaining only overhead court shots and discarding close-up shots) and joint annotation by three experienced experts (who label independently first and reach a consensus through discussion for divergences). The resulting possession-level annotations are stored in a structured JSON format (right), providing precise video-time-anchored labels for visual understanding tasks.}
   \label{fig2}
\end{figure*}
\section{Related Work}
This chapter will respectively introduce mainstream basketball video datasets and related visual understanding tasks.
\subsection{Survey of Basketball Video Datasets}
Researchers have developed several basketball video datasets, each equipped with specific annotations to support their corresponding downstream tasks. The FSN dataset\cite{FSN} is designed for video captioning, consisting of 2,000 high-definition NBA videos and detailed descriptive paragraphs.The NBA dataset\cite{NBA} targets group activity recognition, covering 9 categories of group actions, and adopts video-level labels to reduce annotation costs.The NSVA dataset\cite{NSVA} is the only existing multi-task sports video analysis dataset, supporting video captioning, action recognition, and player identification. It contains 32,019 NBA clips, 172 fine-grained action categories and the identity information of 184 players.The BH-Commentary dataset\cite{BH} is proposed for basketball highlight commentary generation, including 4,396 NBA highlight videos and corresponding professional commentaries.The VC-2022 dataset\cite{xi2025simple} is a player-aware video captioning dataset with 3,977 clips and associated visual and name knowledge of 286 players.The FineSports dataset\cite{fine} focuses on fine-grained spatiotemporal action localization and recognition, containing 10,000 NBA videos and 123,000 annotated spatiotemporal bounding boxes.

Among the mentioned basketball video datasets, except for the NSVA dataset\cite{NSVA} which contains rich annotation information to support more than one downstream task, all other datasets can only cater to a single visual understanding task. To address this issue, our proposed PL-NBA dataset can be adapted to diverse downstream tasks. Table~\ref{tab1} presents the comparison between PL-NBA and aforementioned datasets.

\subsection{Visual Understanding Tasks}
This section presents four typical visual understanding tasks in sports video analysis. It also introduces three novel tasks developed based on the PL-NB dataset.

\noindent \textbf{Event Recognition (ER)}: This task aims to identify the specific event categories related to sports competitions from videos\cite{AR5,AR6}. It takes a video as input and outputs the corresponding event type.

\noindent \textbf{Temporal Action Localization (TAL)}: This task locates the temporal boundaries where actions occur\cite{tal1,anctionformer,tridet}. 

\noindent \textbf{Spatiotemporal Action Localization (SAL)}: This task extends TAL to recognize action categories and temporal boundaries while localizing players via bounding boxes \cite{fine}. PL-NBA provides no bounding box annotations, making it incompatible with SAL tasks.

\noindent \textbf{Video Captioning(VC)}: It aims to convert video content into human-understandable text\cite{goal,soccernet,vc123,clip4vc}.

\noindent \textbf{Action Anticipation}: The proposed task takes pre-offensive historical video clips to predict offensive tendencies: perimeter offense (\textit{2ptShot}, \textit{3ptShot}) and inside offense (\textit{Layup}, \textit{Dunk}). PL-NBA with complete possession videos is the first dataset supporting this task.

\noindent \textbf{Fine-grained Event Recognition}: A novel task derived from PL-NBA, which aims to detect all contained events within a possession. Owing to page constraints, the preliminary experiment results are available at \url{https://github.com/holhouse/PL-NBA-Dataset}.

\noindent \textbf{Fine-grained Video Captioning}:  A novel task based on PL-NBA to generate textual descriptions for complete offensive possessions. Preliminary results will be released via the above link.

\section{Dataset Construction}
This work constructs a possession-level basketball video dataset (PL-NBA). Centered on complete offensive possessions from NBA games, the dataset provides high-quality data support for visual understanding tasks through multiple and refined annotations. The dataset construction pipeline is illustrated in Figure~\ref{fig2}.

\subsection{Data Collection}
Considering the scenario complexity and authority of official NBA games, all samples are derived from 2022–2025 season replays, with 60 official matches selected to cover all 30 league teams.

\noindent \textbf{Perspective control}: Only overhead court shots are retained, as this perspective can fully present player positions, tactical movements, and ball possession transitions. Close-up shots, audience shots, and non-game-related content (e.g., post-game interviews, commercial breaks) are excluded to reduce irrelevant visual noise.

\noindent \textbf{Multi-expert annotation}: Three basketball-dataset annotators label all samples independently. Discrepancies are discussed until a consensus is reached.

\noindent \textbf{Possession Boundary Definition}: A possession starts from the first overhead view when the offensive team gains ball possession, typically with an inbound pass; it ends once the ball possession switches, such as after an offensive score.

\subsection{Data Annotation}
A possession-level video sample contains multiple sub-events. Each sub-event is accompanied by comprehensive annotation, including event type, player name, event outcome, timestamp, and anonymized text description, as shown in the JSON example shown in Figure~\ref{fig2}. 

There are 13 categories of sub-event types, as illustrated in Table~\ref{tab2}. Player names are cross-verified with textual game records from the official NBA website via jersey numbers, manually validated to ensure accuracy. Timestamp labels record the start and end time points of each sub-event in the original video file, with the unit of second:frame (sexagesimal). Event outcome labels are a newly introduced annotation type in the PL-NBA dataset, representing the final result of each sub-event. The corresponding outcomes for different event types are also presented in Table~\ref{tab2}. Anonymized text descriptions fully depict the event content in textual form, generated by Gemini 3 large language model(LLM).

\begin{table}[htbp]
\centering
\caption{Distribution of Sub-Events in PL-NBA Dataset.}
\label{tab2}
\begin{tabular}{lcc}
\hline
\textbf{Sub-Event Type} & \textbf{Sub-Event Outcome} &\textbf{Count} \\
\hline
\textit{Dribble} & \textit{success}          & 8429 \\
 & \textit{fouled}           & 566 \\
 & \textit{turnover}         & 292 \\
     \midrule
\textit{PassBall} & \textit{success}         & 7037 \\
 & \textit{turnover}        & 471 \\
 & \textit{outside}         & 302 \\
     \midrule
\textit{3ptShot} & \textit{made}             & 1005 \\
 & \textit{missed}           & 1641 \\
     \midrule
\textit{2ptShot} & \textit{made}             & 425 \\
 & \textit{missed}           & 495 \\
     \midrule
\textit{Layup} & \textit{made}               & 1131 \\
 & \textit{missed}             & 892 \\
 & \textit{fouled}             & 963 \\
     \midrule
\textit{FreeThrow} & \textit{made}           & 568 \\
 & \textit{missed}         & 295 \\
     \midrule
\textit{Dunk} & \textit{made}                & 614 \\
\textit{Hold} & \textit{success}             & 1933 \\
\textit{HandOff} & \textit{success}          & 800 \\
\textit{InboundPass} & \textit{success}      & 656 \\
\textit{Steal} & \textit{success}            & 685 \\
\textit{DefensiveRebound} & \textit{success} & 1867 \\
\textit{OffensiveRebound} & \textit{success} & 500 \\
\hline
\end{tabular}
\end{table}

\subsection{Dataset Statistics}
PL-NBA contains  11,000 possession-level video samples (average 12.11 s) and 31,567 event-level sub-events, covering 13 sub-event categories and 22 corresponding event types with outcomes. All samples offer synchronized 1080P visuals and 48 kHz audio, enabling commentary extraction and audio-related downstream tasks, like sound event detection. Detailed event distributions are presented in Table~\ref{tab2}. The data follows real-game natural frequencies, forming a long-tailed distribution with frequent events such as \textit{PassBall.success} and rare cases like \textit{Dribble.turnover}. Corresponding strategies are adopted in subsequent experiments to tackle this long-tailed distribution problem.

\subsection{Data Availability Statement}
This dataset is released under the Creative Commons Attribution-NonCommercial 4.0 International License (CC BY-NC 4.0), which explicitly stipulates that the dataset is restricted to academic and research purposes and permits free use, modification, and distribution.

\section{Baseline Methods}
To thoroughly evaluate the effectiveness of the PL-NBA dataset, we conduct benchmark experiments on four visual understanding tasks: Event Recognition, Video Captioning, Temporal Action Localization, and a newly introduced Action Anticipation task.

\subsection{Data Resampling Strategy}
Basketball events exhibit a long-tailed distribution. For event recognition and video captioning tasks, we down-sample \textit{PassBall.success} and \textit{Dribble.success} to 2,000 samples each, and construct a balanced subset containing 20,101 sub-event samples. For temporal action localization and action anticipation, we focus on the localization and prediction of offensive actions (regardless of whether the shot is made or missed), and build a specialized subset with 4,554 offensive possessions that only covers four core offensive categories: \textit{2ptShot}, \textit{3ptShot}, \textit{Layup}, and \textit{Dunk}.

\subsection{Action Anticipation Baseline}
Action anticipation requires the model to predict the forthcoming offensive tendency (perimeter or inside offense) in advance. To establish a baseline for this novel task, we propose a Transformer-based predictive framework. 

The model takes video clips of three historical sub-events (balancing temporal context and computational efficiency) before the initiation of the offensive action as input, and we first uniformly sample $T$ = 90 frames to form the visual sequence $V = \{v_1, v_2, \dots, v_T\}$. A visual backbone extracts spatial features for each frame, generating the sequence of features $F$:
\begin{equation}
F = \text{Backbone}(V) = \{f_1, f_2, \dots, f_T\}
\end{equation}

To capture the complex temporal dynamics and dependencies within the historical context, $F$ is fed into a Transformer encoder-decoder structure. The output sequence is then projected into the offensive mode space, producing the predicted probability distribution $\hat{y}$ over two categories: \textbf{perimeter offense} and \textbf{inside offense}:
\begin{equation}
\hat{y} = \text{Softmax}(\text{Transformer}(F))
\end{equation}

The model is optimized with the Cross-Entropy loss function between predicted probabilities $\hat{y}$ and ground-truth labels $y_c$:
\begin{equation}
\mathcal{L}_{CE} = -\sum_{c=1}^{C} y_c \log(\hat{y}_c)
\end{equation}
where $C = 2$. Specifically, \textbf{perimeter offense} includes \textit{2ptShot} and \textit{3ptShot}, while \textbf{inside offense} includes \textit{Layup} and \textit{Dunk}.

\begin{table*}[t]
\centering
\caption{The Performance of Event Recognition on the PL-NBA Dataset: Precision(\%), Recall(\%) and F1-score.}
\label{tab3}
\setlength{\tabcolsep}{8pt}
\renewcommand{\arraystretch}{1.1}
\begin{tabular}{lccc lccc}
\hline
\textbf{Sub-event} & \textbf{Precision} & \textbf{Recall} & \textbf{F1-score} 
& \textbf{Sub-event} & \textbf{Precision} & \textbf{Recall} & \textbf{F1-score} \\
\hline
\textit{Dribble.success} & 83.11 & 53.06 & 0.6477 
& \textit{Dunk.made} & 92.31 & 66.06 & 0.7701 \\

\textit{Hold.success} & 70.20 & 81.29 & 0.7534 
& \textit{FreeThrow.made} & 77.10 & 100.00 & 0.8707 \\

\textit{DefensiveRebound.success} & 84.62 & 66.47 & 0.7445 
& \textit{Dribble.fouled} & 29.65 & 50.50 & 0.3738 \\

\textit{PassBall.success} & 55.64 & 70.00 & 0.6200 
& \textit{OffensiveRebound.success} & 31.67 & 42.70 & 0.3636 \\

\textit{3ptShot.missed} & 82.20 & 66.67 & 0.7362 
& \textit{2ptShot.missed} & 43.28 & 65.91 & 0.5225 \\

\textit{Layup.made} & 67.66 & 67.66 & 0.6766 
& \textit{2ptShot.made} & 38.60 & 57.89 & 0.4632 \\

\textit{3ptShot.made} & 66.18 & 76.97 & 0.7117 
& \textit{PassBall.outside} & 55.10 & 50.00 & 0.5243 \\

\textit{Layup.fouled} & 70.59 & 74.27 & 0.7256 
& \textit{FreeThrow.missed} & 92.31 & 45.28 & 0.6076 \\

\textit{Layup.missed} & 60.47 & 49.37 & 0.5436 
& \textit{Dribble.turnover} & 40.00 & 15.38 & 0.2222 \\

\textit{HandOff.success} & 87.93 & 35.92 & 0.5100 
& \textit{PassBall.turnover} & 28.57 & 23.81 & 0.2597 \\

\textit{Steal.success} & 42.86 & 49.18 & 0.4580 
& \textit{InboundPass.success} & 87.18 & 87.18 & 0.8718 \\
\hline
\end{tabular}
\end{table*}

\subsection{Baselines for Existing Tasks}
\label{subsec:baselines}
For the three visual understanding tasks, we employ classic networks to build baselines and evaluate their performance on PL-NBA dataset.
For event recognition, we use ResNet-18 as the backbone to extract visual features of 22 sub-events, and introduce the motion enhancement module from the Detector-Free model\cite{dtfree} to embed dynamic motion information of adjacent frames. The fused spatial-temporal features are then fed into a classification network to recognize the 22 sub-events.
For video captioning, we build a baseline framework based on the CLIP4Caption model\cite{clip4vc}. It extracts visual features of sub-events with CLIP\cite{clip}  and converts the visual features into textual descriptions through an encoder-decoder structure.
For temporal action localization, we adopt the TriDet model\cite{tridet} to construct the baseline. We extract video features of full offensive possessions using I3D\cite{I3D}, and output the offensive action categories and their corresponding temporal boundaries via its network structure.

\subsection{Experimental Settings}
\noindent \textbf{Data Partition}: 
We adopt a 7:3 train-test split for all downstream tasks on PL-NBA. Event recognition and video captioning use 20,101 annotated sub-events as experimental data, which are clipped from full videos based on fine-grained timestamp annotations. Temporal action localization and action anticipation adopt 4,554 complete offensive possessions as experimental data, preserving the complete temporal structure and continuous action context.

\noindent \textbf{Evaluation Metrics}: 
Event recognition employs Precision, Recall and F1-score.
Video captioning adopts BLEU-4, CIDEr, ROUGE-L and METEOR.
Temporal action localization adopts average precision under different IoU thresholds (0.3/0.4/0.5/0.6/0.7). Action anticipation adopts classification accuracy as the evaluation metric to measure the correctness of predicting the two high-level offensive tendencies: perimeter offense and inside offense.

\noindent \textbf{Implementation Details}: 
For fair comparison, all baseline models in Section~\ref{subsec:baselines} adopt the same hyperparameters and settings as documented in their original papers.
The proposed action anticipation framework takes Timesformer\cite{timesformer} as backbones to obtain frame-level visual features, which are then fed into a unified Transformer-based encoder-decoder architecture (FANNTRA)\cite{AA} for context modeling, and outputs the final prediction through a linear classification head.

\subsection{Experimental Result}
\textbf{Event Recognition}: 
Table \ref{tab3} shows the recognition results of each subcategory. Overall, several event categories achieve excellent recognition performance, including \textit{InboundPass.success} (0.8718) and \textit{FreeThrow.made} (0.8707), with F1-scores no less than 0.75. These events generally share common characteristics such as highly distinguishable visual features, standardized motion patterns, and strong scene constraints. Specifically, the \textit{FreeThrow.made} category achieves a recall rate of 1.0, mainly due to the highly regular and fixed pattern of the free-throw scenario and the standardized shooting motion, which facilitates effective feature modeling by the model. Similarly, this also applies to \textit{InboundPass.success}. In contrast, \textit{Hold.success} and \textit{HandOff.success}, as relatively static ball-control events, have limited motion amplitude but relatively stable duration and clear ball-possession relationships. Such temporal stability helps the model learn consistent sequential feature representations. However, some event categories still show poor recognition performance with F1-scores below 0.5. Through analysis, two main factors limiting event recognition performance can be summarized: first, visual feature homogeneity. For example, \textit{Steal.success} (0.4580) and \textit{PassBall.turnover} (0.2597) both appear visually as possession changes. Their core difference lies in the underlying causes (defensive interception versus offensive mistakes) rather than observable motion patterns. Due to the lack of explicit discriminative visual cues and causal modeling, the model is prone to misclassification. Second, the lack of high-level semantic information. \textit{OffensiveRebound.success} (0.3636) and \textit{DefensiveRebound.success} exhibit highly similar motion patterns following missed shots. Their key distinction lies in the ball-possession context and offensive/defensive roles. Since the current visual branch does not explicitly model contextual cues such as possession state or team identity, it remains challenging to distinguish different types of rebounds.

\begin{table}[htbp]
\centering
\caption{The Performance of VC Task Using Clip4Caption.}
\label{tab4}
\begin{tabular}{lccc}
\hline
\textbf{CIDEr} & \textbf{METEOR} & \textbf{Rouge-L} & \textbf{BLEU-4} \\
\hline
134.1 & 28.2 & 54.8 & 32.7 \\
\hline
\end{tabular}
\end{table}

\noindent\textbf{Video Captioning}: 
Table \ref{tab4} presents the quantitative results of video captioning on the Clip4Caption model. The evaluation is conducted using four standard metrics: BLEU-4, ROUGE-L, METEOR, and CIDEr.
BLEU-4 measures n-gram precision, focusing on the fluency and lexical accuracy of the generated text. ROUGE-L evaluates the longest common subsequence, verifying the structural and content consistency. METEOR incorporates synonymy and stemming to assess semantic adequacy, while CIDEr, being consensus-based, is highly sensitive to fine-grained event semantics and measures the consistency with human references.
The strong performance across all metrics indicates that the anonymized text description annotations provided by the PL-NBA dataset can supply strong supervisory signals for the Clip4Caption model and help the model generate textual descriptions. These results validate that the PL-NBA dataset can effectively support the video captioning task and generate coherent, information-rich natural language descriptions for basketball game videos.

\begin{figure*}[t]
  \centering
  \includegraphics[width=1.0\linewidth]{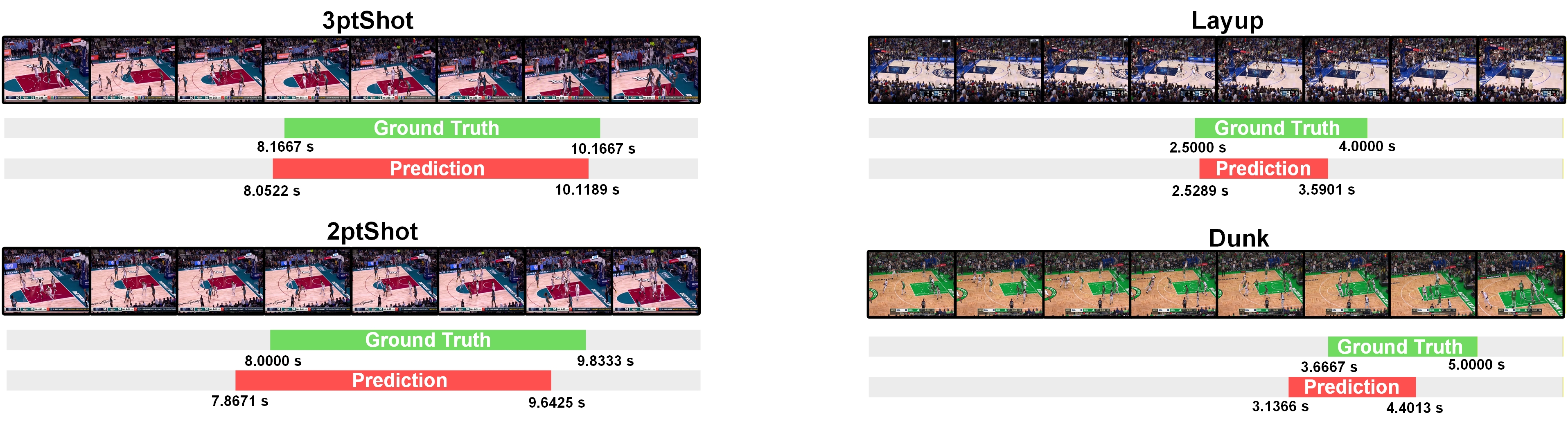}
   \caption{Qualitative temporal action localization results on PL-NBA using Tridet at tIoU=0.5, showing four basketball offensive events: \textit{3ptShot}, \textit{2ptShot}, \textit{Layup}, and \textit{Dunk}. Green bars indicate ground-truth segments, and red bars show model predictions.}
   \label{fig3}
\end{figure*}

\begin{table}[htbp]
  \centering
  \caption{Average Precision (\%) of TAL task Using TriDet.}
  \label{tab5}
  \begin{tabular}{lccccc}
    \hline
    \textbf{Action} / \textbf{tIoU} & \textbf{0.3} & \textbf{0.4} & \textbf{0.5} & \textbf{0.6} & \textbf{0.7} \\
    \midrule
    \textit{3ptShot}  & 91.19 & 91.13 & 89.33 & 85.54 & 79.40 \\
    \textit{2ptShot}  & 51.32 & 50.95 & 50.73 & 44.33 & 35.55 \\
    \textit{Layup}    & 72.67 & 71.37 & 65.93 & 55.63 & 41.37 \\
    \textit{Dunk}     & 25.72 & 23.95 & 19.25 & 13.96 & 7.35 \\
    \midrule
    \textbf{Average}   & \textbf{60.23} & \textbf{59.35} & \textbf{56.31} & \textbf{49.87} & \textbf{40.92} \\
    \hline
  \end{tabular}
\end{table}

\noindent\textbf{Temporal Action Localization}: 
As shown in Table~\ref{tab5}, considerable performance differences are observed among the four offensive events in terms of temporal action localization under various tIoU thresholds using the Tridet model on the PL-NBA dataset. Specifically, \textit{3ptShot} achieves the best localization performance across all tIoU values from 0.3 to 0.7. This is because it occurs in the long-range outer area, with highly distinguishable visual features and clear temporal boundaries, enabling the model to stably capture its unique spatiotemporal characteristics. \textit{2ptShot} and \textit{Layup} achieve moderate detection accuracy. Although they have certain scene recognition features, their performance decreases steadily as the tIoU threshold increases. The poor detection performance of \textit{Dunk} is mainly attributed to two factors: first, it is highly similar to \textit{Layup} visually, with the only critical difference lying in the detailed motion of pressing the ball into the rim at the final stage; second, players usually hang on the basket after completing a dunk, which seriously interferes with the determination of the temporal boundaries of this action. Figure~\ref{fig3} shows the qualitative visualization results of the four offensive events at tIoU=0.5. Notably, the predicted intervals of \textit {3ptShot}, \textit {2ptShot}, and \textit {Layup} are highly consistent with the ground truth, whereas the predicted results of \textit {Dunk} show deviations, aligning with the previous analysis.

\begin{table}[htbp]
  \centering
  \caption{Accuracy(\%) of action anticipation on PL-NBA. }
  \label{tab6}
  \begin{tabular}{lcccc}
    \hline
    \textbf{Action} & \textbf{Fold 1} & \textbf{Fold 2} & \textbf{Fold 3} & \textbf{Mean $\pm$ Std} \\
    \hline
        \textit{3ptShot}          & 73.30 & 74.85 & 70.10 & 72.75 $\pm$ 2.38 \\
    \textit{2ptShot}          & 87.70 & 74.00 & 84.30 & 82.00 $\pm$ 7.20 \\
    \textit{Layup}            & 50.30 & 58.20 & 45.70 & 51.40 $\pm$ 6.31 \\
    \textit{Dunk}             & 63.60 & 69.50 & 57.30 & 63.47 $\pm$ 6.10 \\
                \midrule
    \textbf{Average} & \textbf{66.20} & \textbf{68.72} & \textbf{61.17} & \textbf{65.36 $\pm$ 3.78} \\
    \hline
  \end{tabular}
\end{table}

\noindent\textbf{Action Anticipation}: 
Table~\ref{tab6} reports the three-fold cross-validation results based on random fold splitting for the action anticipation task. This task takes video clips of three historical sub-events before the offensive action as input, and predicts two high-level offensive tendencies: perimeter offense (including 3pt and 2pt shot) and inside offense (including layup and dunk). The proposed baseline achieves an overall accuracy of 65.38±3.7
The experimental results verify that predicting subsequent offensive tendencies based on historical sub-event video observations is feasible. Among the four fine-grained actions, 2ptShot achieves the highest and most stable accuracy of 82.00\%, followed by 3ptShot. In contrast, the low accuracy of Layup degrades the overall prediction performance for inside offense. The current baseline only adopts raw visual and temporal features for modeling, leaving considerable room for improvement. Future work can incorporate fine-grained auxiliary information, such as player bounding box coordinates, court prior knowledge including the three-point line and restricted area, as well as tactical positioning and motion trajectories, to further enhance the model’s ability to distinguish between perimeter and inside offensive tendencies.
\section{Conclusion}
This paper constructs PL-NBA, the first possession-level basketball video dataset capable of supporting various visual understanding tasks. Collected from NBA matches, the PL-NBA dataset contains  11,000 complete offensive possession samples and provides multiple annotations, covering diverse fine-grained events and preserving the contextual relevance among basketball game events. We verify the effectiveness of the dataset on three existing visual understanding tasks: event recognition, video captioning, and temporal action localization. In addition, we propose the action anticipation task for predicting upcoming offensive tendencies with a Transformer-based baseline model. The results confirm the feasibility of conducting tactical prediction based on historical visual context. The fully open-source PL-NBA dataset is expected to offer solid data support for future research in sports visual understanding.








\bibliographystyle{ACM-Reference-Format}
\bibliography{sample-base}










\end{document}